\documentclass[nonatbib]{article}
\usepackage{arydshln}
\usepackage{wrapfig}
\usepackage{etoolbox}
\BeforeBeginEnvironment{wrapfig}{\setlength{\intextsep}{0pt}}

\usepackage{multirow}
\usepackage{comment}
\usepackage{subcaption}
\usepackage{adjustbox}
\usepackage[preprint]{neurips_2023}

\usepackage{booktabs} 
\usepackage{graphicx}
\usepackage{amsfonts}
\usepackage{amsmath}

\def\be{\begin{equation}}
\def\ee{\end{equation}}
\def\bea{\begin{eqnarray}}
\def\eea{\end{eqnarray}}
\usepackage{multicol}
\usepackage[utf8]{inputenc} 
\usepackage[T1]{fontenc}    
\usepackage{hyperref}       
\usepackage{url}            
\usepackage{booktabs}       
\usepackage{amsfonts}       
\usepackage{nicefrac}       
\usepackage{microtype}      
\usepackage{xcolor}         
\usepackage{tikz}
\usetikzlibrary{arrows,snakes,shapes.arrows,decorations.markings}
     \tikzset{>=triangle 90}
     \tikzstyle{bbc}=[draw,circle,fill=black,scale=.75]
     \tikzstyle{rc}=[circle,fill=red,scale=.6]
     \tikzstyle{wc}=[draw,circle,scale=.75]

\def\red#1{{\color{red}{#1}}}

\def\hat{\widehat}

\def\^{\wedge}

\def\g{{\gamma}}

\def\m{{\mu}}

\def\s{{\sigma}}
\def\S{{\Sigma}}

\def\cE{{\mathcal E}}

\def\cP{{\mathcal P}}

\def\beq{\begin{equation}}
\def\eeq{\end{equation}}

\newcommand{\bpmat}{\begin{pmatrix}}
\newcommand{\epmat}{\end{pmatrix}}
\newcommand{\bsmat}{\begin{smallmatrix}}
\newcommand{\esmat}{\end{smallmatrix}}

\title{Taken by Surprise: Contrast effect for Similarity Scores}

\author{%
Thomas C.~Bachlechner$^{1}$\quad Mario Martone$^{1,2}$ \quad Marjorie Schillo$^{1}$\\[5pt]
$^1$EliseAI, New York, NY 10016\\
$^2$King’s College London, London WC2R 2LS, UK\\[5pt]
\texttt{\{thomas,mario,marjorie\}@eliseai.com}
}

\begin{document}

\maketitle

\begin{abstract}
Accurately evaluating the similarity of object vector embeddings is of critical importance for natural language processing, information retrieval and classification tasks. Popular similarity scores (e.g cosine similarity) are based on pairs of embedding vectors and disregard the distribution of the ensemble from which objects are drawn. Human perception of object similarity significantly depends on the context in which the objects appear. In this work we propose the \emph{surprise score}, an ensemble-normalized similarity metric that encapsulates the contrast effect of human perception and significantly improves the classification performance on zero- and few-shot document classification tasks. This score quantifies the surprise to find a given similarity between two elements relative to the pairwise ensemble similarities. We evaluate this metric on zero/few shot classification and clustering tasks and typically find 10-15\% better performance compared to raw cosine similarity.  Our code is available at \href{https://github.com/MeetElise/surprise-similarity}{https://github.com/MeetElise/surprise-similarity}.
\end{abstract} 

\section{Introduction}
Embedding vectors are numerical representations of objects in a high-dimensional vector space \cite{pennington2014glove}. 
These embeddings, which capture the semantic and contextual information in documents, are generated using various algorithms, including deep neural networks. The vector representation allows for efficient evaluation of similarities between objects, which has led to powerful applications for information retrieval, document clustering, recommender systems and summarization \cite{mikolov2013efficient,muennighoff2022mteb,wang2022text,johnson2019billion}. Similar documents generally will have similar vector embeddings.

\begin{wraptable}{r}{5.5cm}
\begin{center}
\begin{tabular}{lrr}
\toprule
Query & $\cos(\text{`dog'},\text{Query})$\\
\midrule
dog & 100.0\\
the & 85.2\\
potato & 85.0\\
my & 85.0  \\
Alsatian & 84.9  \\
furry & 83.3\\
puppyish & 83.0 \\
\bottomrule
\end{tabular} 
  \caption{Cosine similarity ($\times$100) for pairs of words, ranked by cosine score.}
\label{tab:cos-with-dog}
\end{center}
\end{wraptable} 

While there is not a unique measure to quantify the similarity of two objects, where similarity is defined according to human perception, several distance scores operating on embedding vectors have proven successful in practice: Cosine similarity measures the cosine of the angle between two vectors;  Euclidean distance measures straight-line distance between two points in the vector space; Manhattan distance is the sum of the absolute differences between the coordinates of the two points. Each of these scores returns an objective pairwise similarity, disregarding the context within which the objects appear. This poses a problem for classification tasks as no statistically meaningful threshold value can be defined that would indicate an equal similarity across different documents, embedding methods and similarity metrics. Subjective human perception famously adjusts the objective similarity of pairs based on the context. This powerful phenomenon is known as the contrast effect \cite{Sherif1961,Nosofsky1984,shepard1987}: a gray square appears lighter when surrounded by darker squares and darker when surrounded by lighter squares, even though the gray square's actual shade remains the same.

To illustrate the discrepancy between human perception and objective pairwise similarity metrics, we evaluate the cosine scores between embedding vectors for words in the English dictionary\footnote{Our qualitative results are independent of the language embedding model chosen, in this example we use a standard Transformer based model \url{https://huggingface.co/sentence-transformers/sentence-t5-base}. As a simple dictionary we use $25,474$ words contained in \texttt{english\_words\_alpha\_set} shipped with the package \url{https://pypi.org/project/english-words/}.}, and show a number of examples for cosine similarities in Table \ref{tab:cos-with-dog}. A system using the pairwise metric would classify the word `potato' as more closely related to the word `dog' than the word `Alsatian'. 
This may be counter-intuitive since `Alsatian' refers to a specific breed of dog, also known as the German Shepherd, while `potato' is a starchy, tuberous crop and is not related to dogs at all. Despite being out-of-sync with human expectations, this observation can be understood from the origin of the embeddings. During training, a language model captures semantic relationships based on the context in which the words appear. It is possible that in the training data, `potato' and `dog' have appeared together in similar contexts more often than `Alsatian' and `dog'. For example, `potato' and `dog' might appear together in phrases like `hot dog' and `couch potato.' 
In contrast, `Alsatian' is a more specialized term and might not appear as frequently in the training data.

In this work, we propose a new similarity metric that we call the {\it surprise score}: an estimate of the probability that the observed pairwise similarity between a key and a query is higher than the similarity between the query and a randomly chosen element of the ensemble constituting the context (e.g.\,the training data of the underlying embedding model). A given value of the surprise score has an invariant statistical meaning, and therefore allows consistent interpretation across tasks.
Evaluating the surprise score for the `dog' example above by modeling the distribution of cosine scores as a Gaussian distribution (and using the dictionary as the ensemble) we find the surprise scores shown in Table \ref{tab:surprise-dog}. As expected, the word `potato' has a significantly higher typical similarity compared to the word `Alsatian', while both have similar pairwise similarities to the word `dog'. By adjusting the pairwise score based on the context in which the pair appears, the surprise score resembles the contrast effect and more closely matches human perception.

\begin{table}[!t]
\begin{center}
\centering
      \begin{tabular}{lrrrr}
        \toprule
        Query & $\cos(\text{`dog'},\text{Query})$ & mean(cos) & std-dev(cos) &  $\text{surprise}(\text{`dog'},\text{Query})$\\
        \midrule
        dog & 100.0 & 78.5 & 2.3 & 100.0\\
        Alsatian & 84.9 & 77.0 & 2.2 &100.0 \\
        furry & 83.3 & 78.0 & 2.1 &99.5\\
        puppyish & 83.0 & 78.0 & 2.2 &98.8\\
        potato & 85.0 & 80.5 & 2.0&98.7\\
        the & 85.2 & 81.1 & 2.0& 98.0 \\
        my & 85.0 & 80.9 & 2.1&97.4 \\
        \bottomrule
      \end{tabular} 
\end{center}
\caption{Cosine similarity (+ mean \& standard deviations) along with a Gaussian model surprise score for pairs of words, ranked by surprise score.}
\label{tab:surprise-dog}
\end{table}

\paragraph{Related Work} We are unaware of any development of similarity scores that directly account for the context in which the comparison is made. In some sense the semi-supervised learning of the embedding vectors themselves, as pioneered by word2vec \cite{mikolov2013efficient, mikolov2013distributed} and GloVe \cite{pennington2014glove} is the closest approach to incorporating context into vector comparisons.  However, these methods depend on sequences in the embedding-training corpus which may or may not be a relevant context for specific use-cases where we wish to apply similarity evaluations. In the context of document retrieval, there is also some conceptual overlap with TF-IDF and BM25 \cite{robertson1995okapi} systems which adjust the ranking for words that have a high frequency of appearance over many documents in the context, as well as having a loose statistical interpretation \cite{Robertsonarticle}.  The successes of context-awareness in the learning of vector embeddings and document ranking functions provides further motivation for our development of a context-aware pairwise similarity score. 

This work is organized as follows. In Section \ref{sec:Surprise} we formally introduce the surprise similarity score and discuss its general properties. In Section \ref{sec:classification} we apply the surprise score to classification tasks, discussing implementation and results for zero- and few-shots classification. We additionally discuss a new training routine which is effective even for plain cosine similarity. In Section \ref{sec:clustering} we apply the surprise score to document clustering. We finally conclude in Section \ref{sec:discussion} by discussing further applications to clustering and document retrieval and ranking.

\section{Surprise Score}\label{sec:Surprise}
The surprise score $\Sigma(k,q|\mathcal{E})$ is a similarity score between a key $k$ and a query $q$, relative to an ensemble $\mathcal{E}$. Given a set of elements $k, q, e \in \cE$ and a similarity score $\Psi$, the surprise score is defined as the probability for a uniformly sampled member of the ensemble, $e$, to have a similarity to the query, $q$, that is lower than the similarity of the key $k$ to $q$ :
\be
\Sigma(k,q|\mathcal{E})\equiv \mathcal{P}\left(\Psi(e,q)<\Psi(k,q)|e\in \mathcal{E}\right).
\ee

In order to obtain a numerical value for the surprise score, we model the probability distribution of the similarities between each of the queries and the ensemble. Assuming a normal distribution for the similarities across the ensemble, we obtain the approximation
\be\label{surprisescore}
\Sigma(k,q|\mathcal{E}) \approx {1\over 2} \left(1+\text{Erf}\left[{\Psi(k,q)-\m_{\Psi(e,q)|e\in \mathcal{E}}\over \sqrt{2}\sigma_{\Psi(e,q)|e\in \mathcal{E}}}\right]\right)\,,
\ee
where $\sigma_{\Psi(e,q)|e\in \mathcal{E}}$ and $\mu_{\Psi(e,q)|e\in \mathcal{E}}$ are estimates for the standard deviation and mean of the normal distribution of the similarity scores of the query $q$ respectively. For small ensembles, or for non-Gaussian distributions, it can be advantageous to use the p-50 and p-84.14 values instead of the mean and standard deviations, as they are less susceptible to outliers.

While this is a desired property of the contrast effect, it does mean that careful consideration is required in the choice of keys/queries/ensemble for a given problem. For example, in classification tasks, the labels constitute the queries, while the keys and ensemble should represent the documents to be classified. Whereas if the task is to find similar documents within a single document set, the documents constitute all keys, queries as well as the ensemble. For question answering tasks against documents, the ensemble and queries are represented by the questions to be answered while the documents are the keys.

As a probability, the surprise score is bounded $0\le \Sigma\le 1$, and larger values indicate a larger similarity. Typical elements of the ensemble have a surprise score of $\langle\Sigma(e,q)\rangle_\text{median}\approx 0.5$ and since a fixed value of the score has an invariant statistical interpretation, the absolute surprise score for a fixed ensemble is meaningful and can be used across applications. One perhaps counter-intuitive property is that the surprise score is not symmetric:
\be
\Sigma(k,q|\mathcal{E}) \ne \Sigma(q,k|\mathcal{E})\,.
\ee
This is a manifestation of the contrast effect: as a more specialized word, the key `Alsatian' implies the query `dog' more strongly than the key `dog' implies the query `Alsatian,' and therefore the former pairing has a higher surprise score.
While this is a desired property of the contrast effect, it does mean that careful consideration is required in the choice of keys/queries/ensemble for a given problem. For example, in classification tasks, the labels constitute the queries, while the keys and ensemble should represent the documents to be classified. Whereas if the task is to find similar documents within a single document set, the documents constitute all keys, queries as well as the ensemble. For question answering tasks against documents, the ensemble and queries are represented by the questions to be answered while the documents are the keys. 

\subsection{A dynamic crossover towards surprise}
\label{sec:mixed}

The surprise score crucially relies on the statistical model used to estimate the distribution of similarities across the ensemble. If insufficient data is available to accurately estimate the statistical properties of the ensemble, we want to dynamically revert to the plain similarity score. To achieve a smooth interpolation between a pure similarity score and the surprise score we need to carefully deal with the relative normalization of the scores since the surprise score always has a mean of 0.5. To do so, we can define a re-scaled similarity $\hat{\Psi}$ as the similarity $\Psi$ step wise linearly re-scaled to have a mean of $0.5$, while still mapping $\hat{\Psi}|_{\Psi=1}=1$ and $\hat{\Psi}|_{\Psi=0}=0$.

$\hat{\Psi}$ will maintain the original score ordering, but since its mean is $0.5$, just as the surprise score, we can define a mixed surprise score as
\be\label{mixedsurprise}
\Sigma_{w}(k,q|\mathcal{E})\equiv (1-w)\times\hat{\Psi}(k,q)+w\times\Sigma(k,q|\mathcal{E})\,,
\ee
where we define the surprise weight $w$, $0\le w\le 1$. Working under the assumption that increasing the size of the ensemble will allow for more accurate modeling of the context,  we propose an interpolating function based on the cardinality $|\mathcal{E}|$:
\be \label{surprise-weight-Ncross}
\Sigma_{\tanh(|\mathcal{E}|/N_{\text{cross}})}(k,q|\mathcal{E})\,,
\ee
which would return a score ranking based on plain similarity for $|\mathcal{E}|\ll N_{\text{cross}}$, but a surprise score ranking for large ensembles.

\section{Text classification}\label{sec:classification}

To test the effectiveness of the surprise score in classification tasks we perform a series of zero- and few-shot text classification experiments.  For the plain similarity score $\Psi(k,q)$ we choose the cosine similarity between the vector embeddings of $k$ and $q$. For text classification, the correct assignment for the keys, queries and the ensemble is the following:
\beq
k \in \texttt{documents},\quad q \in \texttt{labels}\quad
{\rm and}\quad \cE \equiv \texttt{documents}.
\eeq
where \texttt{documents} corresponds to the set of documents to be classified for each dataset. We make the assumption that the cosine similarity scores between each query and the documents are normally distributed and thus use \eqref{surprisescore} to compute the surprise score. For all classification tasks we map the label given by the dataset to a query via the following:
\be\label{template}
q = \text{``this matter is \textit{label}''}.
\ee

All our experiment are single label classification where we need to identify for each document, which we will denote by $k$, the most relevant label, which we will denote by $q_k$. Using the surprise score, for a set of labels $q\in\mathcal{Q}$, this simply amounts to taking the label with the highest surprise similarity score with $k$:
\be \label{cls_choice}
q_k = \underset{q\in \mathcal{Q}}{\text{argmax}}(\Sigma_{w}(k,q|\mathcal{E}))\,.
\ee

We compare the results we obtain using the surprise score with those of plain cosine similarity. In this case we find the label by maximizing the plain cosine similarity:
\be\label{slabel}
q_k=\underset{q\in \mathcal{Q}}{\text{argmax}}(\Psi(k,q)) = \underset{q\in \mathcal{Q}}{\text{argmax}}(\Sigma_{w=0}(k,q|\mathcal{E}))\,,
\ee
where in the second equality we substituted the mixed surprise score from (\ref{mixedsurprise}). To assess the surprise score performance in a broader context, we choose SetFit, a  popular framework for zero and few-shot classification \cite{2022arXiv220911055T}, as a third point of comparison. The SetFit model  is comprised of two stages: first, fine tune the sentence embeddings using contrastive learning and second, training a logistic regression classification head\footnote{In the zero-shot scenario the first stage is absent, but in order to train the logistic classifier a synthetic dataset is constructed using a default template: `This sentence is \emph{label}'.} which then determines the label chosen for each document. The choice of SetFit is motivated since it is similar to the surprise score in that is lightweight and flexible. Lightweight because, like the surprise score, it can be implemented using any underlying pretrained sentence transformer model. Flexible because, like surprise score, it does not rely on complicated prompt creation and can be applied immediately to multilingual and otherwise varied datasets. 

Since the surprise score takes advantage of the ensemble properties of the chosen similarity score, we expect its performance to be qualitatively independent of the underlying language model chosen to compute embeddings. We have tested this by running the zero-shot classification experiments for three different embedding models: 
\begin{itemize}
    \item {\bf sentence-t5-base}: contains only the encoder of the T5-base model, further pre-trained using contrastive learning on question-answering and natural language inference objectives \cite{ni2021sentencet5}.  This model has 110M parameters. 
    \item {\bf e5-large}: One of the E5, {\bf E}mb{\bf E}ddings from bidir{\bf E}ctional {\bf E}ncoder r{\bf E}presentations, embedding models \cite{wang2022text} that transfer well to a wide range of tasks and which is trained in a contrastive manner with weak supervision on a curated large-scale text pair dataset. This model has 300M parameters.
    \item {\bf gtr-t5-large}: a dual encoder model. At the time of publishing, {\bf G}eneralizable
    {\bf T}5-based dense {\bf R}etrievers (GTR) \cite{DBLP:journals/corr/abs-2112-07899} significantly outperformed
    existing sparse and dense retrievers on the
    BEIR dataset \cite{thakur2021beir}. This model has 335M parameters.
\end{itemize}
Since our results confirm our expectations (see Table \ref{tab:models}), we run the remaining experiments using only the sentence-t5-base model. We describe the specific tasks and datasets in Section \ref{sec:data}. The training details is explained in section \ref{sec:training} and the results are presented in section \ref{sec:results}. 

\subsection{Classification tasks} \label{sec:data}

We benchmark our classification experiments on five standard datasets:
\begin{itemize}
    \item AG News \cite{zhang2015character} contains a dataset of news articles' titles and descriptions paired with a category from the labels: \emph{`Business'}, \emph{`Sci/Tech'}, \emph{`Sports'}, or \emph{`World'}.  The test set is comprised of 7,600 samples evenly distributed among the 4 categories. 
    
    \item Yahoo! Answers \cite{zhang2015character} contains a dataset of questions paired with a topical category such as: \emph{`Business or Finance'},
    \emph{`Computers or Internet'},
    \emph{`Family or Relationships'} or
    \emph{`Science or Mathematics'}.
    The test set is comprised of 60,000 samples evenly distributed among the 10 categories. 

    \item DBPeida \cite{zhang2015character} contains a dataset of DBPedia\footnote{A project whose goal is to extract structured data from Wikipedia \cite{dbpedia}.} titles and abstracts paired with one of 14 ontological categories such as \emph{`Artist'}, \emph{`Company'}, \emph{`NaturalPlace'}, or \emph{`OfficeHolder'}.  The test set is comprised of 70,000 samples evenly distributed among the 14 categories. 
    
    \item Amazon Reviews Multi \cite{marc_reviews} contains a dataset of customer reviews paired with one of 31 retail categories such as: \emph{`apparel'},  \emph{`electronics'}, \emph{`grocery'}, or \emph{`video games'}.  We use only the English-language dataset; the test set contains 5,000 samples but is highly imbalanced across the categories, with a range of 2 to 440 samples per category. 

    \item IMDB \cite{maas-EtAl:2011:ACL-HLT2011} contains a dataset of movie reviews paired with a binary sentiment of \emph{`positive'} or \emph{`negative'}. The test set is comprised of 25,000 reviews evenly balanced between the two labels. 
\end{itemize}
These datasets were chosen to give some representation to small, medium and large label set sizes, to cover both formal and colloquial language, and to showcase balanced and unbalanced test sets.

\begin{table}[t]
\begin{center}
  \begin{adjustbox}{center}
  \small
  \centering
  \setlength{\tabcolsep}{4pt}
      \begin{tabular}{l|rrr|rrr|rrr}
        \toprule
        Dataset & \multicolumn{3}{|c|}{\emph{Sentence-t5-base}}& \multicolumn{3}{c|}{\emph{e5-large}} & \multicolumn{3}{c}{\emph{gtr-t5-large }} \\
        Accuracy Scores & SetFit  & Cosine & Surprise &SetFit  & Cosine & Surprise &SetFit  & Cosine & Surprise \\
        \midrule
        Yahoo Answers& 54.3 & 57.8& \underline{\bf{59.7}} & 47.8 & 53.1& \bf{56.4} & 47.0 & 51.5 & \bf{53.9}\\
        Amazon Reviews & \bf{24.4}& 24.0 & 24.3  & 21.9& 22.4 & \bf{23.1} &
        23.9 & \underline{\bf{25.2}} & 25.0 \\
        AG News  & 71.8 & 71.8 & \underline{\bf{75.0}} & 63.5 & 58.6 & \bf{64.7} &
        57.0 & \bf{68.1} & 67.6        \\
        DBpedia& 71.1 & 71.4 & \bf{77.5} & 66.7 & 75.5 & \underline{\bf{80.3}} &
        60.5 & 60.4 &\bf{67.7}\\
        IMDB &  66.2 & 63.9 & \underline{\bf{83.9}}& 74.5 & 79.6 & \bf{79.9} &
        77.7 & 78.8 &\bf{80.3}\\ \midrule
        \emph{average} & 57.6 & 57.8 & \underline{\bf{64.1}} & 54.9 & 57.8 & \bf{60.9} & 53.2 &56.8 & \bf{58.9} \\ \midrule& & & & & & & & & \\

        F1 Scores & & & & &  &  &  \\
        \midrule
        Yahoo Answers & 52.0& 57.2& \underline{\bf{58.7}} & 46.4& 51.9& \bf{55.3} &
        45.7 & 49.8 &\bf{53.3}\\
        Amazon Reviews & 23.2& 22.2 & \bf{23.8}  & 20.5& 20.7 & \bf{22.8} &
        22.7 & 22.2 & \underline{\bf{24.2}}\\
        AG News  & 69.7 & 70.7 & \underline{\bf{74.9}} & 61.4 & 50.5 & \bf{64.0} &
        56.0 & \bf{67.7} & 67.4\\
        DBpedia    & 68.3 & 69.2 & \bf{76.7}& 66.5 & 74.2 & \underline{\bf{79.6}} &
        58.2 & 55.7 & \bf{66.8}\\
        IMDB & 62.0 & 58.8 &  \underline{\bf{83.9}}& 73.1 & 79.6 &\bf{79.9} &
        77.2 & 78.6 &\bf{80.3}\\  \midrule
        \emph{average} & 55.0 & 54.4 & \underline{\bf{63.6}} & 53.6 & 55.4 & \bf{60.3} & 52.0 & 54.8 & \bf{58.4}\\
        \bottomrule
      \end{tabular}
  \setlength{\tabcolsep}{6pt}
    \end{adjustbox}
  \end{center}
  \caption{Zero-shot classification metrics comparing SetFit \cite{2022arXiv220911055T}, maximum cosine score across labels and maximum surprise score across labels. The underlying pretrained embedding models used for SetFit, cosine scores, and surprise scores are, from left to right, the sentence-t5-base model \cite{ni2021sentencet5}, the e5-large model \cite{wang2022text} and the gtr-t5-large model \cite{DBLP:journals/corr/abs-2112-07899}. In bold the score which perform best for each embedding model, while the best across all models is further underlined.}
  \label{tab:models}
  \vspace{-1em}
\end{table}

\subsection{Training} 
\label{sec:training}

For the few-shot classification we use the following new training routine which takes inspiration from a contrastive approach \cite{Koch2015SiameseNN} using in addition a mixed binary focal loss objective. For each key to be classified, $k$, we form pairs $(k,q)$ with target queries modeled as in \eqref{template} over all the labels in the dataset. 
The pair is designated as a \emph{positive pair} if the target query is associated to $k$'s correct label, while  all remaining pairs are designated as \emph{negative pairs}. We associate different target weights $i_{(k,q)}$ to positive and negative pairs with:
\beq\label{pairsw} 
i_{(k,q)}:\qquad\left\{
\begin{array}{cl}
1 & {\rm positive\ pairs}\\
\epsilon & {\rm negative\ pairs}.
\end{array}
\right.
\eeq
We find that the use of a non-zero weight $\epsilon$ for negative pairs dramatically improves performances on datasets with large set of labels.  The reason behind this is that in the case of a large label set, negative pairs dominate the training sample giving a high likelihood of entire batches without positive pairs; a small non-zero epsilon prevents the model from over-reacting to this imbalance and skewing towards negative predictions. While we did not observe much sensitivity to the precise value, we fix a default value to be $\epsilon=0.05$.

\paragraph{Focal loss} \hspace{-10pt} As a loss-function we use a binary focal loss. The focal loss FL$(p_{(k,q)})$ \cite{2017arXiv170802002L} is a variation of the cross-entropy criterion\footnote{Note that the standard definition uses weights, $i$, either zero or one, but we have generalized to include the modification \eqref{pairsw}.} CE$(p_{(k,q)}, i_{(k,q)})=-i_{(k,q)}\log{p_{(k,q)}} - (1-i_{(k,q)})\log(1-p_{(k,q)})$, where $p_{(k,q)}\in [0,1]$ is the model estimated probability for $k$ to be labled as $q$.   FL$(p_{(k,q)})$ is defined as:
\beq\label{focal}
{\rm FL}(p_{(k,q)}, i_{(k,q)}):=(1-p_{(k,q)})^\g\, {\rm CE}(p_{(k,q)},i_{(k,q)}).
\eeq
The focal loss reduces the relative loss for well-classified examples ($p_{(k,q)} > 0.5)$ thus \emph{focusing} on hard, misclassified ones. We fix a default value of $\g$ in \eqref{focal} to be $\g=1$. To fine-tune the underlying sentence transformer model, we also fix the learning rate to be $0.0001$ and the weight decay to be 0.01 .

Further, $p_{(k,q)}$ is computed by the cosine similarity of the $k$ and $q$'s vector embeddings, which we denote $\Psi(k, q)$. To make sure that the outcome is a probability estimate we apply a ReLU to the cosine similarity adding a very small off-set $\delta$ (a value of $10^{-10}$ suffices) to avoid instabilities in the CE criterion:
\beq
p_{(k,q)}:=\s_{\rm ReLU}\big(\Psi(k, q)+\delta\big).
\eeq 
All in all our loss-function is:
\be
\ell(k,q):=-i_{(k,q)}\big(1-p_{(k,q)}\big)\log\big(p_{(k,q)}\big)-\big(1-i_{(k,q)}\big)p_{(k,q)}\log\big(1-p_{(k,q)}\big).
\ee

\paragraph{Convergence} In lieu of fixing a number of training epochs hyperparameter, we use a threshold on the cross-entropy criterion averaged over number of batches per epoch to determine when training has converged.  This serves as a more universal stopping criterion which applies well to a wide variety of training dataset-sizes.  We fix the threshold for stopping training to be when $\langle CE \rangle_{batch} < .3$.

\subsection{Results} \label{sec:results}

\begin{figure*}[t]
    \centering
    \begin{subfigure}[t]{0.45\textwidth}
        \centering
        \includegraphics[width=.85\textwidth]{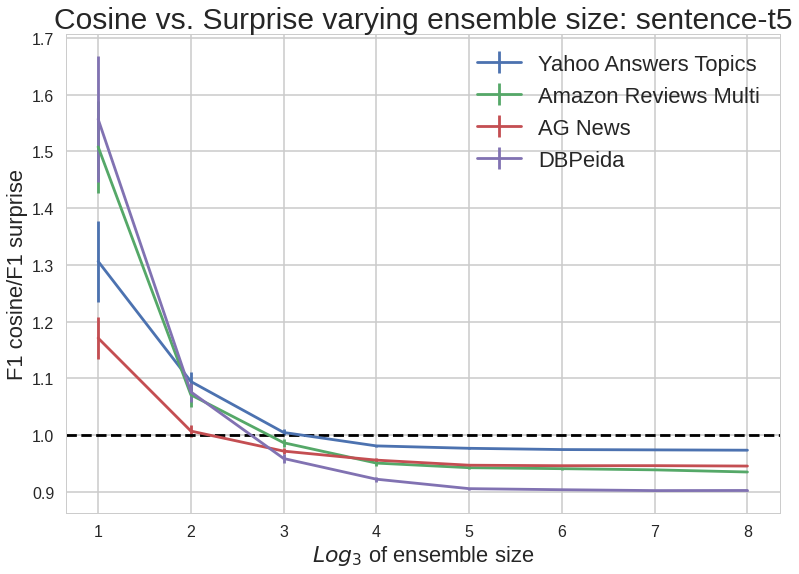}
    \end{subfigure}%
    ~ 
    \begin{subfigure}[t]{0.45\textwidth}
        \centering
        \includegraphics[width=.95\textwidth]{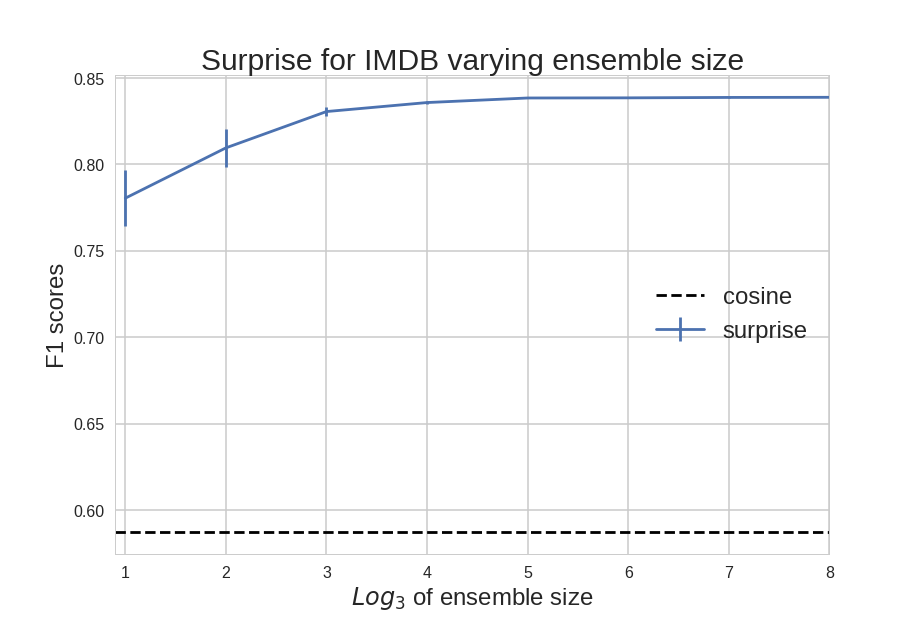}
    \end{subfigure}
    \caption{(a) Ratio of zero-shot F1 scores using cosine similarity over using surprise similarity with varying ensemble size. (b) Zero-shot F1 scores for each similarity score with varying ensemble size. \label{fig:ensemble}}
\end{figure*}

\paragraph{Zero-shot} In Table \ref{tab:models} we report zero-shot F1 and accuracy scores for the datasets discussed in Section \ref{sec:data}. The results are similar across various language models and, in nearly all cases, the surprise score achieves the best performance, sometimes by a very large margin. 

\paragraph{Cosine vs. Surprise varying the ensemble} We analyse the effectiveness of the surprise score as a function of the ensemble size. To do so, we look at the various datasets discussed in Section \ref{sec:data}. We compute F1 scores of zero-shot classifications using a mixed surprise weight \eqref{mixedsurprise} evaluated both at $w=0$ (pure cosine similarity) and $w=1$ (pure surprise similarity). The latter is evaluated on an ensemble of increasing size randomly sampled out of the entire test set. We plot the ratio of the average F1 score (cosine/surprise) in Figure \ref{fig:ensemble}. A ratio greater than one clearly indicates that the cosine similarity is more accurate than surprise and viceversa. We choose eight benchmark ensemble sizes:
\beq\label{benchensemble}
{\rm ensemble\ sizes:}\quad\{3,9,27,81,243,729, 2187\}
\eeq
which correspond to $3^n$ for $n=1,...,7$, thus justifying our choice to plot the $\log_3\{$ensemble size$\}$ in Figure \ref{fig:ensemble}. For each value of the ensemble size in \eqref{benchensemble} we make ten runs where we choose the subset of the test set at random. We use these runs to compute the averages and 1-standard deviation error bars. 

As can be seen from the figure, as the size of the ensemble grows, surprise similarity becomes considerably more accurate than a plain cosine similarity with a roughly `universal' cross-over point of $|\mathcal{E}|\sim100$\footnote{For the \emph{IMBD} dataset the surprise score always outperforms cosine as it is shown in Figure \ref{fig:ensemble}(b).}.

\paragraph{Few-shots} For our few-shot experiments, all run using the sentence-t5-base embedding model, we engineer both a balanced and an unbalanced training dataset from the original training sample of  each of the five datasets discussed above. In the balanced case, we build our training dataset by picking at random $k$ samples per label, with $k=3,6,9,12,15,18,21$. In the unbalanced case, we pick at random a subset of the training sample of size proportional to $n_{\rm label}$, the total number of labels in the dataset. Specifically  we pick the size to be $k\cdot n_{\rm label}$, with $k=3,6,9,12,15,18,21$ and $n_{\rm label}=4, 31, 14, 2$, and 10 for AG News, Amazon Reviews Multi, DBPedia, IMDB, and Yahoo Answer Topics respectively. To give statistical significance to our findings we perform ten runs for each value of $k$. We then compare the F1 scores obtained by fine-tuning a surprise score classifier and a cosine score classifier both trained via the procedure in Section \ref{sec:training}, as well as a SetFit model \cite{2022arXiv220911055T} using all default parameters.  In computing our prediction for the surprise score, we use the mixed surprise score \eqref{surprise-weight-Ncross} with $N_{cross}=1000$. This in particular implies that the surprise weight we use in our experiments is effectively equal to 1 since the test sizes are all larger than 1000.

\begin{center}
\begin{figure*}[t]
    \begin{subfigure}[t]{\textwidth}
        \centering
        \includegraphics[width=1\textwidth]{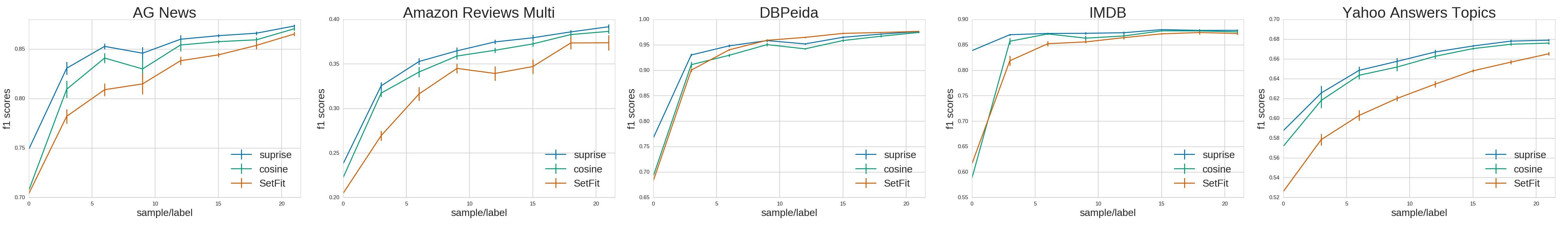}
        \caption{Few-shots results with a balanced training dataset. 
  \label{fig:balanced}}
    \end{subfigure}%
    \\
    \begin{subfigure}[t]{\textwidth}        \includegraphics[width=1.04\textwidth]{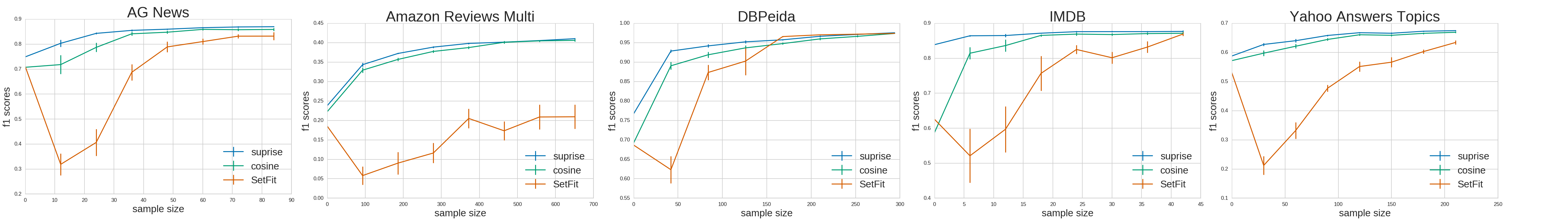}
        \centering
        \caption{Few-shots results with an unbalanced training dataset.\label{fig:unbalanced}}
    \end{subfigure}
    \caption{(a) F1 score as a function of the number of training sample per label. (b) F1 score as a function of the training sample size. The errors are computed over ten runs per value of each data point. From left to right we report the results for the AG News, Amazon Reveiws Multi, DBPedia, IMDB and Yahoo Answer Topics dataset. \label{fig:few-shots}}
\end{figure*}
\end{center}

For a balanced training set, the results can be seen in Figure \ref{fig:balanced}. The plots are obtained by interpolating among the aforementioned values of $k$ and the error represent the one standard deviation from the mean. The surprise similarity score is consistently better for $k\leq9$. In the case of the DBPedia and IMDB dataset cosine and SetFit catch up for larger $k$, whereas for the other datasets surprise remains superior to cosine even for larger training sample, with SetFit performing significantly worse.

The results for unbalanced few-shot training sets are instead reported in Figure \ref{fig:unbalanced}. Here the results are even more pronounced with the surprise similarity score outperforming SetFit by a large margin for all datasets at all training set sizes, except for DBPedia, where performance is similar for large datasets.  That the cosine similarity also outperforms SetFit provides favorable evidence for the effectiveness of our training routine (see Section \ref{sec:training}) in addition to the use of the surprise score.

\section{Clustering} \label{sec:clustering}
K-means is an unsupervised algorithm used for clustering data points based on their similarity. Given an ensemble and a number of clusters $k$, the algorithm first identifies $k$ centroid vectors and then assigns all elements in the ensemble to the centroid with highest similarity. The simplest application of the surprise score to clustering is to identify the centroids in the traditional way (e.g. by using cosine similarity) but to employ the surprise score to determine cluster assignment of all elements.  In this application the queries are the set of centroids, and the elements to be clustered make up both the keys and the ensemble. 

To evaluate the relative performance of the cosine similarity to the surprise similarity score we perform clustering on the 25 test splits of the RedditClustering dataset \cite{geigle2021tweac,muennighoff2022mteb}. We use a vanilla K-Means algorithm\footnote{With K-means++ initialization \cite{kmeanspp}} to determine the centroids of embeddings generated from the sentence-t5-base model. For each of the 25 splits, we repeat clustering 40 times and evaluate one entropy based clustering metric (v-measure score) and one pairwise agreement score (the adjusted rand score)\footnote{The qualitative results we find hold true for other entropy based metrics such as the mutual information score, and other pairwise agreement metrics such as the Fowlkes-Mallows score, but we will not report the quantitative results for brevity.} and report the results in Table \ref{tab:clustering}.
\begin{table}[t]
  \small
  \centering
  \setlength{\tabcolsep}{4pt}
      \begin{tabular}{lrr}
        \toprule
        Metric & Cosine Score & Surprise Score\\
        \midrule
        V-measure & $54.2\pm 1.8$&$54.4\pm 1.9$\\
        Adjusted Rand Score & $39.5\pm 1.8$& $ 44.3\pm 1.9$\\
        \bottomrule
      \end{tabular}
  \setlength{\tabcolsep}{6pt}
  \caption{Entropy (v-measure) and pairwise correctness (Adjusted Rand Score) metrics for the clustering performance of cosine similarity and surprise similarity scores.}
  \label{tab:clustering}
  \vspace{-1em}
\end{table}

The v-measure is an entropy-based measure that evaluates clustering by considering both homogeneity and completeness. Homogeneity means that each cluster should contain only members of a single class, while completeness means that all members of a given class should be assigned to the same cluster. The adjusted rand score is a measure based on pairwise agreement between the true labels and the predicted assignments, adjusted for chance. It counts the number of pairs that are either in the same group in both the true labels and the predicted clusters or in different groups in both. 
The surprise similarity score performs better than cosine similarity when a good agreement between predicted and true labels is required, while it performs on par with cosine similarity when evaluating the clustering on homogeneity and completeness.

\section{Discussion} \label{sec:discussion}

In this  manuscript we have introduced the \emph{surprise score}, a new similarity score which takes into account the context in which objects appear. 
This use of context provides a score that more accurately reproduces the subjective human notion of similarity mimicking the contrast effect in psychology. 
The surprise similarity score is modeled over a pairwise plain similarity score and it measures the probability that the given similarity between a key and a query is higher than the similarity between the query and a randomly chosen element of the ensemble constituting the context. 

We have performed a variety of experiments choosing as a plain similarity score the cosine similarity between objects' vector embeddings and have compared the performance of the surprise score against that of plain cosine. In the case of zero- and few-shots text classification, we have also compared the performances of surprise with SetFit, a popular classification model which relies on contrastive learning and a logistic classification head. We have also compare performances across multiple embedding models. Our results show that the surprise score consistently outperforms everything else and, particularly for some text classification tasks, by a large margin. The training routine that we use for our few-shots experiments is also new.

The surprise score is an extremely flexible similarity score that can be applied beyond what discussed in this work. One example in the context of document ranking and retrieval. In this case, one might adopt different strategies depending on whether the target is part of the same distribution as the documents or not. In the former case we can use the documents themselves as the ensemble $\mathcal{E}$, compute the surprise similarity score among them and then sort the documents accordingly.
If instead the target \emph{is not} part of the same distribution as the documents (\emph{e.g.}\,the BEIR dataset \cite{thakur2021beir}), the targets constitute the keys and ensemble, while the documents are identified as the queries.
In the case of the BIER dataset, the ensemble statistical properties are highly unbalanced with both large variations of relevant documents per target and/or a few documents considerably more relevant than the rest across targets. We expect this to have a negative effect on the surprise score performance, but leave this investigation to future research.

Finally, in this work we have only studied the surprise similarity score modeled over cosine similarity. It would be interesting to analyze whether the prominent advantage acquired by using surprise over cosine persists for other plain similarity scores such as the Euclidean or Manhattan distances.

\bibliographystyle{unsrt}


\appendix

\end{document}